R E S C I E N C E  C

# [Re] CLRNet: Cross Layer Refinement Network for Lane Detection

Viswesh N[1], Kaushal Jadhav[1], Avi Amalanshu[1], Bratin Mondal[1], Sabaris Waran[1], Om Sadhwani[1], Apoorv Kumar[1] Debashish Chakravarty[1]
[1]IIT Kharagpur

## Reproducibility Summary

**Scope of Reproducibility** – The following work is a reproducibility report for CLRNet: Cross Layer Refinement Network for Lane Detection [1]. The basic code was made available by the author at this https url. The paper proposes a novel Cross Layer Refinement Network to utilize both high and low level features for lane detection. The authors assert that the proposed technique sets the new state-of-the-art on three lane-detection benchmarks.

**Methodology** – The proposed model employs a two-stage approach to lane detection. Initially, coarse lane detection is achieved through the extraction of high-level semantic features. This is followed by refinement of the output based on low-level features, aimed at enhancing the localization accuracy of the model. The authors' code was used to benchmark the claims. Some further experiments were investigated thereafter. Kaggle, a free-to-use platform for deep learning experiments, was used to train these models. We have reproduced the code base in Pytorch Lightning and found consistent results across the board.

**Results** – The central claims presented by the authors were subject to reproduction and verification. The validity of the claims was evaluated using two out of the three datasets referenced in the original paper. The results obtained from the CULane dataset showed close agreement with the original findings, with deviations of less than 1% on most of the metrics. This suggests the reproducibility and reliability of the claims made by the authors. However, in experiments on the TuSimple dataset, substantial disparities were noted between our results and those reported in the original paper. The probable causes of these inconsistencies are discussed in the study.

**What was easy** – Obtaining the proposed results on the CULane dataset was readily achievable. The codebase provided by the authors was well-documented and functional. Owing to the modularity of the code, further experiments could be run with minimal changes overall. Porting the codebase to PyTorch Lightning was also facile.

**What was difficult** – Using the LLAMAS dataset proved to be a challenge for resource constrained students owing to its size. We were eventually unable to set up experimentation on that dataset. Limited computational resources proved to be a challenge even for the other datasets, with each epoch taking over 2 hours on CULane. Total training time exceeded a day on the CULane datasets for each individual backbone, while TuSimple was trained for about 10 hours on each backbone.

**Communication with original authors** – Attempts were made to contact the authors over email regarding the obtained results and the discrepancies in the TuSimple benchmark. Unfortunately, we were unable to receive a response from the authors' end, and proceeded with the reproduction to the best of our understanding.

# 1 Introduction

The reproduced paper aims to improve the efficacy of lane detection by utilizing existing techniques as a foundation. Previous lane detection approaches have relied heavily on parameter-based techniques and pre-defined features, such as anchor points and segmentation maps. In contrast, the network introduced in the paper classifies features into two broad categories: high-level and low-level.

The proposed method, Cross Layer Refinement Network (CLRNet), involves detecting high-level semantic features to coarsely localize lanes, followed by refinement with the help of fine-detail features to achieve more precise localization. The progressive refinement of the lane location and feature extraction boosts the accuracy and robustness of the model.

Often, there is no visual evidence of lanes available. To address this challenge, the model establishes a relationship between RoI lane features and the entire feature map, using the novel ROIGather module . The authors claim this helps capture significant global contextual information. Additionally, the novel Line IoU (LIoU) loss is defined to regress the lane as a whole, instead of considering sample points. The authors claim that training the network using this loss results in a significant improvement in performance compared to the standard smooth-l1 loss.

To reproduce the authors' experiments and perform our own, we have used the original codebase . We visualized some new parameters and searched for better hyperparameters. Further, we have carried out an ablation study by varying the weight of each loss component. These experiments are explained in detail in the following sections.

# 2 Scope of reproducibility

Lane detection is a mature problem in computer vision critical to unlocking perception and vehicular intelligence in context of autonomous navigation. The problem that this paper addresses is detecting lanes across various lighting and weather conditions. The central claims of the paper can be summarized as follows:

- High Level and Low level features: The paper discusses the dichotomy of features in lane detection and proposes a novel network, CLRNet, to fully exploit it. The authors claim this leverage allows the model to produce state-of-the-art results in robustness and accuracy, measured by F1 score, accuracy, and false classification rates.

- ROIGather: The importance of global context in lane detection has been incorporated through ROIGather. In settings where the lane instance is occupied or blurred with extreme lighting conditions, the ROIGather module proves to be resourceful. The method is an extension of Mask R-CNN[2] and essentially involves using lane priors to obtain enhanced features. This is followed by further convolution steps to gather context of nearby pixels. The exact details have been elaborated upon in Section 3.

- Line IoU loss: A new evaluation metric and loss function specifically tailored for lane detection is introduced. The Line IoU metric, in contrast to the traditional Intersection over Union (IoU) metric, takes into account the regression of the entire lane rather than a restricted subset of points. To assess its effectiveness, an ablation study is conducted on the Line IoU loss.

## 3 Methodology

The base model was adopted from the author's publicly available GitHub repository. However, to replicate the additional experiments, modifications were required as the repository only included the base model's results on various datasets.

### 3.1 Model descriptions

The network architecture consists of a ResNet [3] backbone with Feature Pyramid Network (FPN)[4] and the ROIGather module. Figure 1 shows the architecture of CLRNet. The Cross Layer Refinement starts from the highest level layer ($L_0$) and gradually approaches the lowest level layer ($L_2$), using a sequence of refinements $\{R_0, R_1, R_2\}$ to refine the lane priors
The ROIGather module takes the feature map and lane prior as input, and uses bilinear interpolation to gather information of nearby pixels for each point in the lane prior. This helps reinforce occupied parts and allows the network to better exploit contextual information to learn better feature representations.
Finally, a Line IOU loss has been defined and the datapoints are regressed using the LIoU loss for training. The following sections elaborate further on the lane priors, ROIGather module and the regression steps.

**Lane Representation –** In traditional object detection methods, objects are represented using rectangular boxes which are not suitable for representing elongated objects such as lane lines. To address this issue, a lane is represented using a sequence of 2D points, denoted as
$$P = (x_1, y_1), \cdots, (x_N, y_N)$$
The y coordinate of points is equally sampled through image vertically i.e. $y_i = \frac{H}{N-1} * i$ where H is the image height and x coordinate is associated with the respective $y_i$. The lane priors proposed in the paper are this obtained
The lane prior is composed of four key components: (1) foreground and background probabilities, (2) the length of the lane prior, (3) the starting point of the lane line and the angle between the x-axis of the lane prior, and (4) the N offsets, representing the horizontal distance between the prediction and its ground truth.

**Cross Layer Refinement –** The backbone architectures utilized in this study are ResNet and DLA34, and the feature levels generated by the Feature Pyramid Network (FPN) are denoted as $\{L_0, L_1, L_2\}$. Cross layer refinement begins with the highest level $L_0$ and gradually proceeds towards $L_2$, and the corresponding refinements are represented as $\{R_0, R_1, R_2\}$. A sequence of refinements is built through the operation
$$P_t = P_{t-1} \circ R_t (L_{t-1}, P_{t-1}),$$
where $t = 1, \cdots, T$, T is the total number of refinements. This method performs detection by first utilizing the highest level layer with high semantic information. The parameter of lane prior, $P_t$, includes the start point coordinates $(x, y)$ and angle $\theta$, and is learnable. For the first layer $L_0$, the $P_0$ is uniformly distributed on the image plane. The refinement $R_t$ takes $P_t$ as input to extract the ROI lane features and subsequently performs two fully connected (FC) layers to obtain the refined parameter $P_t$. The progressive refinement of the lane prior and feature extraction is crucial for the success of the cross layer refinement method. It should be noted that this method is not limited to the FPN structure and is also suitable for use with just the ResNet architecture or the Progressive Attention Feature Pyramid Network (PAFPN) [5].

**ROIGather –** The ROIGather module takes the feature map and lane priors as inputs, where each lane prior has N points. For each lane prior, the ROIAlign operation is performed to obtain the ROI feature of the lane prior $\left(\mathcal{X}_p \in \mathbb{R}^{C \times N_p}\right)$. Unlike the ROIAlign used for bounding boxes, $N_p$ points are sampled from the lane prior and the input features are computed using bilinear interpolation at these locations. To enhance feature representations, the ROI features from previous layers ($L_1$, $L_2$) are concatenated with the ROI features of the current layer. Convolution is then performed on the extracted ROI features to gather nearby features for each lane pixel. To save memory, a fully-connected operation is applied to further extract the lane prior feature $\left(\mathcal{X}_p \in \mathbb{R}^{C \times 1}\right)$.

To gather the global context for the features of lane priors, the attention matrix $\mathcal{W}$ is computed between the ROI lane prior feature ($\mathcal{X}_p$) and the global feature map map ($\mathcal{X}_f$), using the equation:

$$\mathcal{W} = f\left(\frac{\mathcal{X}_p^T \mathcal{X}_f}{\sqrt{C}}\right)$$

where f is the normalization function softmax. The aggregated feature is then calculated as

$$\mathcal{G} = \mathcal{W}\mathcal{X}_f^T$$

The output $\mathcal{G}$ reflects the contribution of $\mathcal{X}_f$ to $\mathcal{X}_p$, which is selected from all locations of $\mathcal{X}_f$. Finally, the output is added to the original input $\mathcal{X}_p$.

**Line IoU Loss –** The Line Intersection over Union (LIoU) loss introduced in the paper extends the concept of IoU to line segments. Each point in the predicted lane is first extended with a radius $e$ (a tunable hyperparameter) into a line segment. The IoU is then computed between extended line segment and its ground truth as follows:

$$\text{IoU} = \frac{d_i^o}{d_i^u} = \frac{min(x_i^p+e, x_i^g+e) - max(x_i^p-e, x_i^g-e)}{max(x_i^p+e, x_i^g+e) - min(x_i^p-e, x_i^g-e)}$$

where $x_i^p + e, x_i^p - e$ are the extended points of $x_i^p$, $x_i^g + e, x_i^g - e$ are the corresponding ground truth points.
The LIoU can be considered as the combination of infinite line points. To simplify the expression and make it easy to compute we discretize it as follows:,

$$LIoU = \frac{\sum_{i=1}^N d_i^{\mathcal{O}}}{\sum_{i=1}^N d_i^u}$$

The LIoU loss is then finally defined as

$$\text{Loss} = 1 - LIoU$$

The Line IoU value ranges from -1 to 1, where a value of 1 indicates perfect overlap between the two line segments, and a value of -1 indicates no overlap. Additionally, the LIoU loss has the advantages of being simple and differentiable, enabling efficient parallel computation, and providing a more holistic prediction of the lane.

## 3.2 Datasets

The paper claimed to drastically outperform other state-of-the-art approaches on three major datasets: CULane [6] , TuSimple and LLAMAS [7]. We have performed extensive experimentation with two of these datasets and the results have been summarized below:

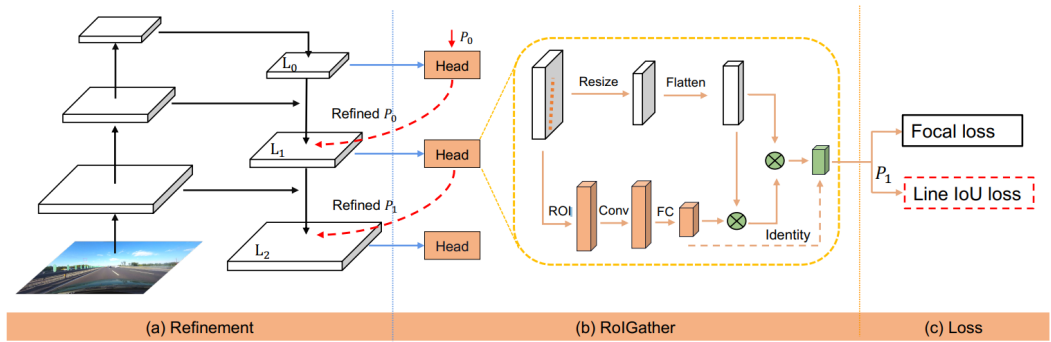

**Figure 1**. Figure 1: Graphical Representation of the CLRNet architecture

**CULane —** The Culane dataset consists of 88880 images for training, 9675 images for validation, and 34680 images for testing. The original image dimensions of 1640x590 have been resized to 320x800. Data augmentation techniques such as random affine transformation and random horizontal flipping were applied during the preprocessing phase. The training process was conducted for a total of 15 epochs using the Culane dataset. The test set is comprised of nine categories, including normal and eight challenging categories such as crowded, night, no line, shadow, arrow, dazzle light, curve, and crossroad. The dataset can be obtained from here

**TuSimple —** The TuSimple dataset consists of 6,408 images which have been split into 3,626 images for the training split, 358 images for the validation split, and 2,782 images for the test split. The original resolution of the images is 1280x720 and was resized to 320x800 for the purpose of training. Data augmentation techniques, including rotation, translation, scaling, and random horizontal flips, were applied to the TuSimple dataset. The training process was conducted for 70 epochs using the preprocessed dataset. The processed TuSimple dataset can be accessed from the follwing link.

**LLAMAS —** The LLAMAS dataset is a large-scale dataset with over 100,000 images, totaling over 100GB. We were therefore unable to replicate the results reported in the original paper due to limitations in connectivity and computational resources.

## 3.3 Hyperparameters

| Hyper-parameters | Value |
| --- | --- |
| initial learning rate | 1e-03 |
| number points of lane prior (N) | 72 |
| sampled number (Np) | 36 |
| $w_{cls}$ | 1 |
| $w_{sim}$ | 3 |
| resized H in ROI Gather | 10 |
| resized W in ROI Gather | 25 |
| channel C | 64 |
| cosine decay learning rate power | 0.9 |
| extended radius (e) in LIoU | 15 |

**Table 1**. Hyperparameters used all over the experiments.

The default hyperparameters described in the paper were employed in our reproduction efforts and are presented in Table 1. In addition, a comprehensive hyperparameter search was carried out beyond the parameters proposed in the paper, and the results are detailed in Section 4.2.

### 3.4 Experimental setup and code

The complete code both using PyTorch [8] and PyTorch Lightning [9] along with WandB [10] integration is available at these links: (PyTorch, PyTorch Lightning). The code for this experiment is primarily established in the main.py file, which accepts the path to the configuration file and GPU number as its arguments. Replicating the pipeline strcutre of [11] and [12], we set up Interactive Python Notebooks so as to facilitate the setup and training processes. Detailed instructions about how to download the dataset and required libraries and setup the training and evaluation processes are given in the ReadMe section of the repository.

The primary metric of evaluation utilized in this study is the F1@50 score, which is calculated by computing the intersection-over-union (IoU) between the predicted lanes and the ground truth. The predicted lanes that exhibit an IoU greater than a specified threshold are deemed as true positive (TP) results. The mathematical formulation of these metrics have been provided below:

Precision:

$$\frac{\text{Number of true positives}}{\text{Number of true positives + Number of false positives}}$$

Recall:

$$\frac{\text{Number of true positives}}{\text{Number of true positives + Number of false negatives}}$$

F1 score:

$$\frac{2*\text{Precision}*\text{Recall}}{\text{Precision + Recall}}$$

### 3.5 Computational requirements

All the experiments have been performed on Kaggle cloud service platform with NVIDIA Tesla P100-PCIE-16GB (NVIDIA-SMI 470.82.01, Driver Version: 470.82.01, CUDA Version: 11.4). The time required for various experiments is mentioned in Table 2.

| Dataset  | Backbone  | Time (in hours) |
|----------|-----------|-----------------|
| CULane   | Resnet34  | 30              |
| CULane   | Resnet18  | 30              |
| CULane   | Resnet101 | 30              |
| CULane   | DLA34     | 30              |
| TuSimple | Resnet34  | 8.5             |
| TuSimple | Resnet18  | 8.5             |
| TuSimple | ResNet-101| 8.5             |

**Table 2.** Approximate time required for training each backbone.

# 4 Results

We experimented and verified all the central claims made by the paper about Cross Layer Refinement and ROIGather structure on CULane dataset and TuSimple dataset. Following are the detailed description of the results obtained

## 4.1 Results reproducing original paper

**Experiments on Tusimple dataset** – We successfully replicated the results reported in Table 3 from the original paper, and it supports the claim of achieving a new state-of-the-art in terms of F1 score on TuSimple.

| Backbone | F1(%) | | Accuracy(%) | | False Positive(%) | | False Negative(%) | |
|---|---|---|---|---|---|---|---|---|
| | paper | ours | paper | ours | paper | ours | paper | ours |
| ResNet18 | 97.89 | 97.73 | 96.84 | 96.75 | 2.28 | 2.27 | 1.92 | 2.27 |
| ResNet34 | 97.82 | 97.76 | 96.87 | 96.81 | 2.27 | 2.37 | 2.08 | 2.07 |
| ResNet101 | 97.62 | 97.75 | 96.83 | 96.65 | 2.37 | 1.69 | 2.38 | 2.81 |

**Table 3**. Comparison of F1 score, Accuracy, False Positive, False Negative on Tusimple dataset

**Experiments on CULane dataset** – We successfully replicated the results reported in Table 4 and Table 5 from the original paper, and it supports the claim of achieving a new state-of-the-art on CULane dataset

| Backbone | mF1 | | F1@50 | | F1@75 | |
|---|---|---|---|---|---|---|
| | paper | ours | paper | ours | paper | ours |
| ResNet-18 | 55.23 | 54.47 | 79.58 | 78.9 | 62.21 | 61.58 |
| Resnet-34 | 55.14 | 54.24 | 79.73 | 78.99 | 62.11 | 61.16 |
| Resnet-101 | 55.55 | 55.26 | 80.13 | 79.64 | 62.96 | 62.69 |
| DLA34 | 55.64 | 55.45 | 80.47 | 79.68 | 62.78 | 62.77 |

**Table 4**. Comparison of mF1 score, F1@50 and F1@75 on CULane dataset

| Backbone | Category-wise (F1@50) | | | | | | | | | | | | | | | | Cross (FP) | |
|---|---|---|---|---|---|---|---|---|---|---|---|---|---|---|---|---|---|---|
| | Normal | | Crowded | | Dazzle | | Shadow | | No line | | Arrow | | Curve | | Night | | | |
| | paper | ours | paper | ours | paper | ours | paper | ours | paper | ours | paper | ours | paper | ours | paper | ours | paper | ours |
| ResNet-18 | 93.3 | 93.2 | 78.33 | 77.68 | 73.71 | 74.56 | 79.66 | 75.8 | 53.14 | 52.11 | 90.25 | 89.66 | 71.56 | 69.3 | 75.11 | 73.72 | 1321 | 1249 |
| ResNet-34 | 93.49 | 93.38 | 78.06 | 77.26 | 74.57 | 72.49 | 79.92 | 81.06 | 54.01 | 53.01 | 90.59 | 90.02 | 72.77 | 71.42 | 75.02 | 74.33 | 1216 | 1385 |
| ResNet-101 | 93.85 | 93.49 | 78.78 | 78.07 | 72.49 | 72.63 | 82.33 | 80.39 | 54.5 | 54.53 | 89.79 | 89.41 | 75.57 | 72.66 | 75.51 | 74.97 | 1262 | 1276 |
| DLA34 | 93.73 | 93.48 | 79.59 | 77.95 | 75.3 | 74.67 | 82.51 | 80.35 | 54.58 | 53.67 | 90.62 | 89.78 | 74.13 | 72.85 | 75.37 | 75.1 | 1155 | 1202 |

**Table 5**. Comparison of Category-wise F1@50 score and Cross(FP) on CULane dataset

## 4.2 Results beyond original paper

**Computation of Category-Wise F1@75 score on Culane dataset** – We additionally calculated F1@75 score for Culane dataset for each of the category. We observed that the F1@75 scores for all the categories are lower than the corresponding F1@50 scores. We find that the scores follow a similar trend as of F1@50 scores.

| Backbone | Category-wise (F1@75) | | | | | | | | Cross (FP) |
|---|---|---|---|---|---|---|---|---|---|
| | Normal | Crowded | Dazzle | Shadow | No line | Arrow | Curve | Night | |
| ResNet-18 | 0.78 | 0.6 | 0.48 | 0.51 | 0.36 | 0.75 | 0.42 | 0.54 | 1249 |
| Resnet-34 | 0.78 | 0.58 | 0.5 | 0.59 | 0.37 | 0.75 | 0.39 | 0.53 | 1385 |
| Resnet-101 | 0.79 | 0.6 | 0.5 | 0.57 | 0.38 | 0.75 | 0.45 | 0.54 | 1276 |
| DLA34 | 0.79 | 0.6 | 0.51 | 0.56 | 0.38 | 0.75 | 0.42 | 0.54 | 1202 |

Table 6. Category-wise F1@75 score on Culane dataset

**Experiments with $W_{cls}$ and $W_{xytl}$ –** We executed ablation studies by changing values of $W_{cls}$ and $W_{xytl}$ in the model on DLA34 backbone and computed the corresponding scores.

| Experiment | mF1 | | F1@50 | | F1@75 | |
|---|---|---|---|---|---|---|
| | Obtained | Paper | Obtained | Paper | Obtained | Paper |
| Wcls=0 | 5.34 | 55.64 | 14.42 | 80.47 | 4.21 | 62.78 |
| Wcls=0, Wxytl=0.5 | 1.75 | 55.64 | 5.48 | 80.47 | 0.64 | 62.78 |
| Wcls=0, Wxytl=1.0 | 19.29 | 55.64 | 33.1 | 80.47 | 19.86 | 62.78 |
| Wcls=0, Wxytl=1.5 | 1.07 | 55.64 | 2.09 | 80.47 | 0.98 | 62.78 |

Table 7. Comparison of F1 scores

| Experiment | Category-wise F1@50 | | | | | | | | | | | | | | | | | |
|---|---|---|---|---|---|---|---|---|---|---|---|---|---|---|---|---|---|---|
| | Normal | | Crowded | | Dazzle | | Shadow | | No line | | Arrow | | Curve | | Night | | Cross(FP) | |
| | Obtained | Paper | Obtained | Paper | Obtained | Paper | Obtained | Paper | Obtained | Paper | Obtained | Paper | Obtained | Paper | Obtained | Paper | Obtained | Paper |
| Wcls=0 | 19.03 | 93.73 | 15.56 | 79.59 | 1.25 | 75.3 | 17.49 | 82.51 | 10.04 | 54.58 | 15.75 | 90.62 | 19.86 | 74.13 | 12.13 | 75.37 | 121150 | 1155 |
| Wcls=0, Wxytl=0.5 | 6.61 | 93.73 | 7.24 | 79.59 | 4.9 | 75.3 | 4.18 | 82.51 | 2.79 | 54.58 | 6.23 | 90.62 | 5.59 | 74.13 | 4.85 | 75.37 | 12488 | 1155 |
| Wcls=0, Wxytl=1.0 | 41.4 | 93.73 | 36.87 | 79.59 | 18.35 | 75.3 | 32.53 | 82.51 | 17.49 | 54.58 | 37.91 | 90.62 | 32.39 | 74.13 | 34.98 | 75.37 | 12487 | 1155 |
| Wcls=0, Wxytl=1.5 | 4.35 | 93.73 | 2.02 | 79.59 | 1.25 | 75.3 | 0.37 | 82.51 | 1.05 | 54.58 | 3.31 | 90.62 | 1.91 | 74.13 | 0.31 | 75.37 | 11980 | 1155 |

Table 8. Comparison of Category-wise F1@50 score on CULane dataset

**Experiments with Radius –** We executed ablation studies on radius(e) of the model

| IoU | F1 score | Tp | Fp | Fn | precision | Recall |
|---|---|---|---|---|---|---|
| 0.5 | 0.7993 | 77054 | 10853 | 27832 | 0.876539 | 0.73464 |
| 0.55 | 0.78159 | 75343 | 12564 | 29543 | 0.857076 | 0.7183 |
| 0.6 | 0.75929 | 73193 | 14714 | 31693 | 0.8326 | 0.69783 |
| 0.65 | 0.7292 | 70297 | 17610 | 34589 | 0.7996 | 0.67022 |
| 0.7 | 0.6873 | 66257 | 21650 | 38629 | 0.7537 | 0.6317 |
| 0.75 | 0.627719 | 60510 | 27397 | 44376 | 0.68834 | 0.57691 |
| 0.8 | 0.537135 | 51778 | 36129 | 53108 | 0.589 | 0.49365 |
| 0.85 | 0.4005 | 38615 | 49292 | 66271 | 0.439271 | 0.36816 |
| 0.9 | 0.202517 | 19522 | 68385 | 85364 | 0.222 | 0.1861 |
| 0.95 | 0.0183 | 1765 | 86142 | 103121 | 0.02007 | 0.0168 |
| mean | 0.5543 | 534334 | 344736 | 514526 | 0.6078 | 0.50944 |

Table 9. Results obtained after changing the extended radius (e) to 10

| IoU | F1 score | Tp | Fp | Fn | Precision | Recall |
|---|---|---|---|---|---|---|
| 0.5 | 0.79937 | 77768 | 11919 | 27118 | 0.8671 | 0.7414 |
| 0.55 | 0.7813 | 76011 | 13676 | 28875 | 0.8475 | 0.7247 |
| 0.6 | 0.75819 | 73762 | 15925 | 31124 | 0.82243 | 0.70325 |
| 0.65 | 0.7271 | 70741 | 18946 | 34145 | 0.7887 | 0.6744 |
| 0.7 | 0.6838 | 66525 | 23162 | 38361 | 0.74174 | 0.63426 |
| 0.75 | 0.62292 | 60602 | 29085 | 44284 | 0.6757 | 0.577789 |
| 0.8 | 0.53416 | 51967 | 37720 | 52919 | 0.579426 | 0.49546 |
| 0.85 | 0.39596 | 38522 | 51165 | 66364 | 0.4295 | 0.36727 |
| 0.9 | 0.2003 | 19489 | 70198 | 85397 | 0.2173 | 0.1858 |
| 0.95 | 0.018 | 1758 | 87929 | 103128 | 0.0196 | 0.01676 |
| mean | 0.5521 | 537145 | 359725 | 511715 | 0.5989 | 0.512 |

Table 10. Results obtained after changing the extended radius (e) to 20

**Experiments with LIoU** – We modified the LIoU function to

$$L_{IoU} = \frac{\sum_{i=1}^{N}(d_i^{\mathcal{O}})^2}{\sum_{i=1}^{N}(d_i^u)^2}$$

in the DLA34 backbone and computed the following scores.

| IoU | F1 score | Tp | Fp | Fn | Precision | Recall |
|---|---|---|---|---|---|---|
| 0.5 | 0.795 | 77405 | 12369 | 27481 | 0.86222 | 0.73799 |
| 0.55 | 0.777396 | 75664 | 14110 | 29222 | 0.8428 | 0.72139 |
| 0.6 | 0.75496 | 73481 | 16293 | 31405 | 0.8185 | 0.700579 |
| 0.65 | 0.72454 | 70520 | 19254 | 34366 | 0.78552 | 0.672349 |
| 0.7 | 0.68293 | 66470 | 23304 | 38416 | 0.74041 | 0.63373 |
| 0.75 | 0.6207 | 60414 | 29360 | 44472 | 0.672956 | 0.57599 |
| 0.8 | 0.53195 | 51775 | 37999 | 53111 | 0.576726 | 0.4936 |
| 0.85 | 0.39339 | 38289 | 51485 | 66597 | 0.4265 | 0.365 |
| 0.9 | 0.2005 | 19515 | 70259 | 85371 | 0.21737 | 0.18605 |
| 0.95 | 0.016849 | 1640 | 88134 | 103246 | 0.018268 | 0.015636 |
| mean | 0.549854 | 535173 | 362567 | 513687 | 0.59613 | 0.51024 |

Table 11. Results obtained after modifying the LIoU function

| Category | F1@50 | F1@75 |
|---|---|---|
| Normal | 0.933 | 0.7877 |
| Crowd | 0.7779 | 0.5987 |
| Hlight | 0.7312 | 0.5069 |
| Shadow | 0.7939 | 0.537 |
| Noline | 0.535 | 0.37326 |
| Arrow | 0.899739 | 0.749 |
| Curve | 0.71357 | 0.37077 |
| Cross | 1112 | 1112 |
| night | 0.7514 | 0.542919 |

Table 12. Category wise scores obtained after modifying the LIoU function

**Ablation studies on non-maximum Suppression** – The original methodology uses non-maximum Suppression to mitigate the effect of highly overlapping lanes. During training, each ground truth lane is assigned with top-k predicted lanes dynamically as positive samples. We set top-k=1 and do a one-to-one assignment for the ground truth lane from the set of predicted lanes. This modification makes our approach nms free. We train the model on the CULane dataset for 15 epochs with DLA34 as the backbone.

| Experiment | mF1 | | F1@50 | | F1@75 | |
|---|---|---|---|---|---|---|
| | Obtained | Paper | Obtained | Paper | Obtained | Paper |
| top-k=1 | 21.34 | 55.64 | 40.73 | 80.47 | 23.05 | 62.78 |

Table 13. F1 scores obtained after setting top-k=1

| Experiment | Category-wise F1@50 | | | | | | | | | | | | | | | | | |
|---|---|---|---|---|---|---|---|---|---|---|---|---|---|---|---|---|---|---|
| | Normal | | Crowded | | Dazzle | | Shadow | | No line | | Arrow | | Curve | | Night | | Cross(FP) | |
| | Obtained | Paper | Obtained | Paper | Obtained | Paper | Obtained | Paper | Obtained | Paper | Obtained | Paper | Obtained | Paper | Obtained | Paper | Obtained | Paper |
| top-k=1 | 44.32 | 93.73 | 41.09 | 79.59 | 40.422 | 75.3 | 44.39 | 82.51 | 33.52 | 54.58 | 41.45 | 90.62 | 40.28 | 74.13 | 39.45 | 75.37 | 2262 | 1155 |

Table 14. Category-wise scores obtained after setting top-k=1

## 5 Visualizations

The following section presents visualizations of the segmented lane markings under various environmental conditions. As shown in Figure 2, several noteworthy observations can be made. The model displays good capability in handling adverse lighting conditions, such as low light and over-exposure, where the lane markings are difficult to discern. However, the last two cases exhibit the model's limitations in detecting curved lane markings

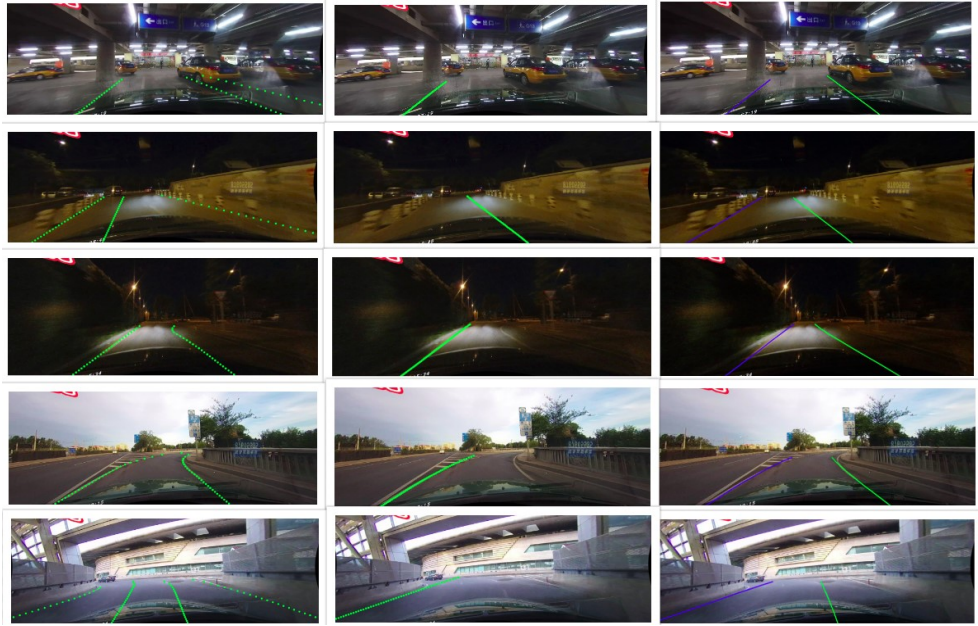

**Figure 2.** Figure 2: Visualization of detected lane markings. The progression from left to right encompasses the following components: the ground truth labels, the intermediate predicted points on the lane outputs generated by the network and the final output of the network.

## 6 Discussion

Through our experiments, we reproduce and verify the central claim of the paper that utilizing both the high level and low level features improvises the accuracy of lane detection.The importance of various terms in the loss function used in the paper gets highlighted through our ablation studies on the loss function. The results we obtain are consistent with the results obtained by the authors.

### 6.1 Methodological Experiments

**Model Performance**: We observe that the F1 scores calculated at different IoU thresholds show a decreasing behavior on increasing the threshold. We argue that even though F1@50 score is significantly greater than F1@95 score, F1@95 metric captures the performance of the model under better localization scenarios. Hence, the mF1 metric is a better metric as it accounts for the performance of the model in all the localization scenarios.

**Focal Loss**: The results obtained upon dropping the focal loss component $w_{cls}$ of the loss were substantially different, to the point where training was vacuous. This justifies the large contribution of the weight to the overall loss function as proposed in the paper. Focal loss [13] helps to focus on the examples that have a very high chance to be classified correctly. In lane detection, focal loss prioritizes the lane points that are easy to be detected and prevents easily misclassified points from affecting the network. Therefore, the focal loss is pivotal for training.

**Non-Maximal Suppression**: We conducted experiments with and without non-maximal suppression (NMS). We found NMS to be important to assign low priority to highly overlapping lanes. However, as discussed in [14], for real-time applications where time savings are crucial, the detector can be made NMS-free if no significant metric drop is observed. We observe a 30% drop in performance when the detector is NMS-free. This is significant for any application, therefore the detector should always use NMS.

### 6.2 Experiments with $L_{IoU}$ loss

**Sum of squares Loss Function**: The value of $d_i^{\mathcal{O}}$ in the $L_{IoU}$ loss equation can be negative. The authors claim that negative values of $d_i^{\mathcal{O}}$ facilitate optimization in scenarios with non-overlapping lanes. In this case, the ground truth and the predicted lanes are distanced at $2e$ from each other, where $e$ is the extended radius. To investigate the effect of constraining the term $d_i^{\mathcal{O}}$ to be strictly positive, we changed the summation in equation $L_{IoU}$ to sum of squares. We found that the metric values obtained are within close range of the original metric scores. Upon further investigation, we observed that the modified loss function displays behavior similar to the original loss function as seen in Figures 3 and 4. Hence we argue that the model weight optimization curve remains unchanged even when the non-overlapping line segments are penalized with a positive value.

**Effect of Radius**: We also analyzed the effect of extended radius on the $L_{IoU}$ loss. We observed that the metric values obtained upon changing the value of radius to e=10 and e=20 display negligible deviations from the value of 15 that has been mentioned in the paper. However, we note that significant deviations from the above mentioned value will lead to the line segments extending out of the region of interest, thus causing a drop in performance.

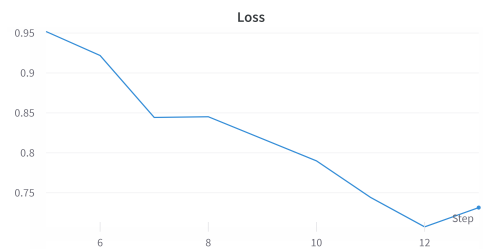

**Figure 3.** Figure 3: Loss curve obtained using the original loss function

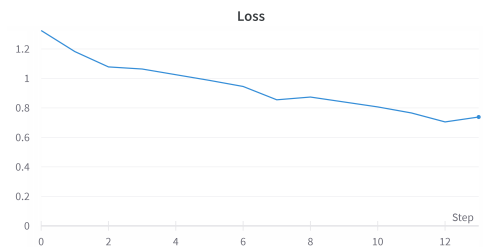

**Figure 4.** Figure 4: Loss curve obtained using the modified loss function

### 6.3 Robustness towards Adverse Conditions

Lane detection in adverse conditions is a challenging task but also an important one for the feasibility of level-5 autonomous vehicles. The CuLane dataset contains test images split into various categories based on common adverse conditions (Crowded, Dazzle, Shadow, No line, Arrow, Curve, and Night). We further present a brief discussion on the comparison of the model's performance on these categories and also discuss countermeasures to tackle them.

**No Line**: We observe that the No Line category has the lowest F1 score. Although poor performance is expected in this category, the F1 scores are still more than half those of the Normal category. This can be attributed to the newly introduced RoI Gather method that establishes a relationship between the ROI lane feature and the entire feature map. This enables a better representation of lane features even when there is no visual evidence of lanes in the image.

**Night** and **Shadow**: We also note that the performance drop in the case of Night and Shadow categories is due to low illumination conditions. Although night segmentation is extremely challenging, the performance in low-light conditions can be improved by adapting the model to low-illumination settings. On the contrary, bright white patches present in the images of the Dazzle category affect the performance of the network. Here, the utilization of coarsely located high-level features helps in the detection of lane features as the highly localized bright patches do not affect the global lane structure.

**Arrow**: The presence of arrow-like features and their similarity with lane-like features is the reason for the drop in performance on the Arrow category. However, the ROIGather Structure prevents F1 scores from dropping very low by classifying arrow features as no-lane features. To improve performance, an additional network devoted to differentiating between lane markings and path guidance markings can be introduced in the network architecture.

**Curve**: Curved roads pose a challenge to lane detection algorithms due to non-linear features. In CLRNet, 2D points are equally sampled along the height of the lane and used as lane priors. The above methodology fails for roads with sharp curves. The model initially starts off with straight-line predictions and then continues to regress the lane as a whole unit making use of the individual point distances . This would be expected to work for curved lanes too, but the model struggles to regress a curved lane from straight line priors. As curves are highly non-linear, bilinear interpolation does not work well for curvy roads. Instead, we propose to include spline interpolated priors in addition to the linearly interpolated lanes.

## 6.4 Reproducibility of the codebase

After porting the code base into PyTorch Lightning, we validated the correctness of training and evaluation processes such as data loading and preprocessing, initializing and optimizing model weights, calculating loss according to the loss function, and saving and reloading checkpoints for further reproducibility experiments in the future.

## 6.5 Future work

The proposed architecture fails frequently in cases where the lanes are curved. To address this, a possible solution could be to use spline interpolation while sampling points for lane priors in place of linear interpolation, so as to not constrain the obtained lane priors to a straight line. Implementing attention based models to capture poor lighting and combining it with the network could also be investigated. Inclusion of spline interpolated lane priors additionally could bring diversity about the nature of lane shapes.

# 7 Conclusion

In this study, we attempt to validate the claims and results of the original research paper on two of the three datasets. Our evaluations affirm the first claim of the original paper, which emphasizes the distinction between high-level and low-level features, as our results show state-of-the-art performance on the CULane dataset. The proposed network effectively integrates information from both levels of features, supporting the first claim of the original paper. Our experiments also confirm the second claim of the original paper regarding the ROIGather module. The module effectively integrates context information in cases where lane markings are absent, supporting the original claim. Our results on the CULane and TuSimple datasets validate the third and final claim of the original paper, concerning the LIoU loss function. By modifying the cost function, we further demonstrate the versatility of this novel loss function. However, it should be noted that the mathematical rationale for the selection of the loss function remains inadequate, as indicated by our experiments with a revised cost function. Furthermore, the results obtained with the TuSimple dataset was not state-of-the-art as proposed in the paper. The limitations of the proposed method include a lack of generalizability to curved lane markings and frequent failure in low-light conditions.

# Appendix

## 8 Wandb Training Logs

### 8.1 Training on TuSimple

The following training logs document the results obtained during the training of CLRNet using Resnet18, Resnet34, and Resnet101 architectures on the TuSimple dataset, using the parameters specified by the authors.

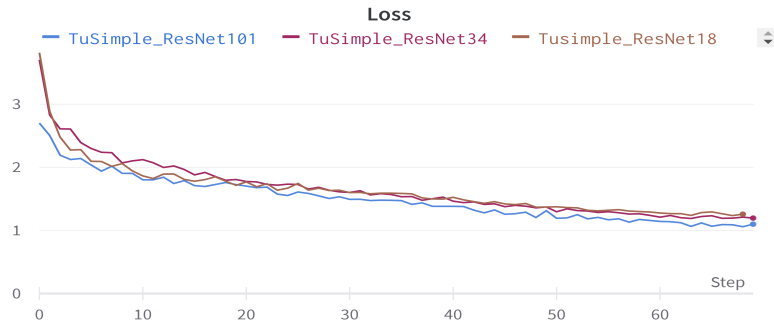

**Figure 5.** Training Loss

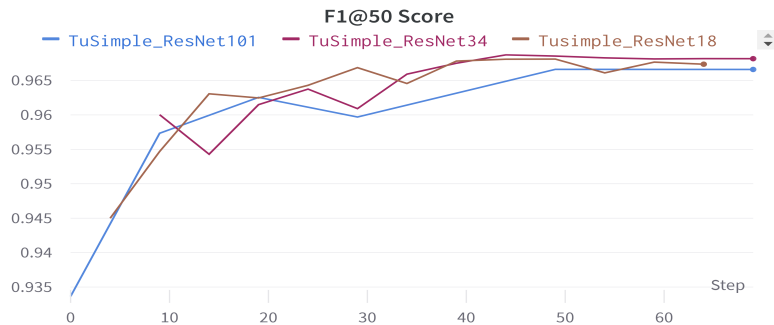

**Figure 6.** F1@50 Scores during training

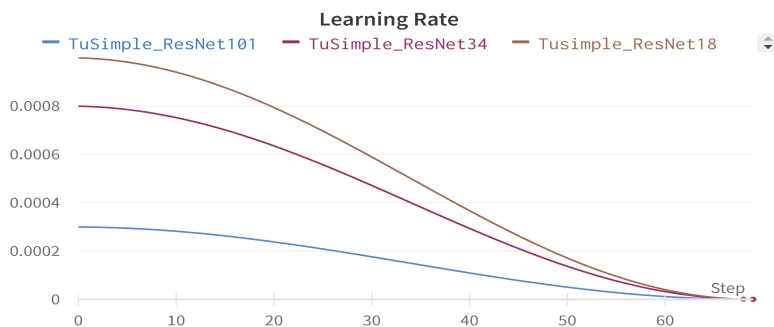

**Figure 7.** Learning Rate during training

### 8.2 Experiments with Extended Radius in LIoU

The following training logs depict the results obtained during the experimentation with an extended radius of 10 and 20 in the loss function.

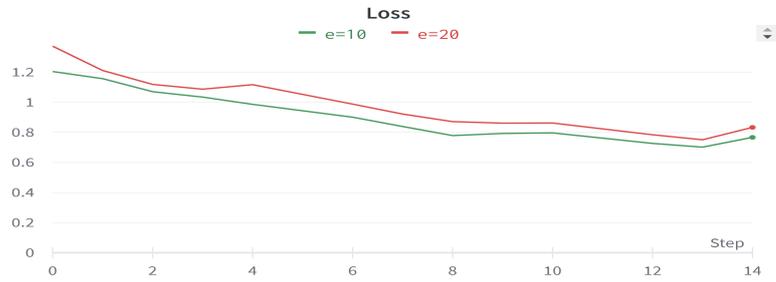

Figure 8. Training Loss

### 8.3 Experiments with NMS

The following training logs present the results obtained during the experimentation with a non-maximum suppression (NMS) parameter of top-k = 1, resulting in a NMS-free model.

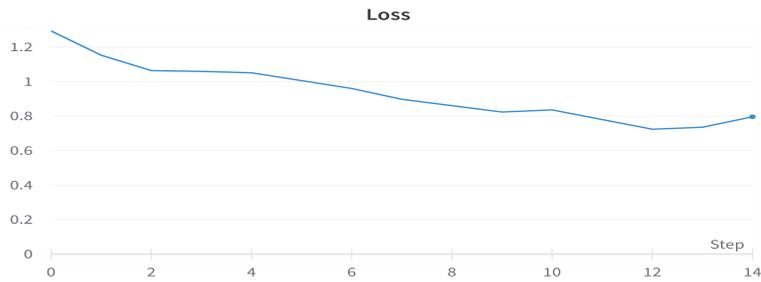

Figure 9. Training Loss

### 8.4 Experiments with modifying weights in loss function

The following training logs depict the results obtained during experimentation with varying weights for $W_{cls}$ and $W_{xytl}$.

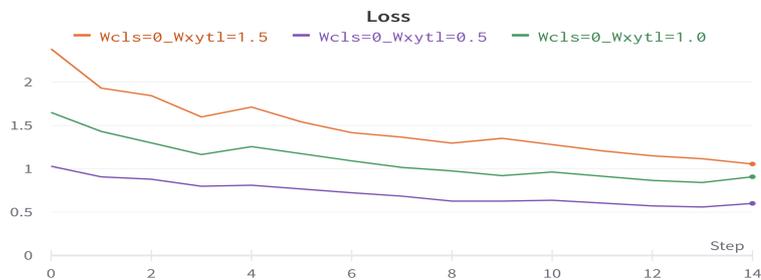

Figure 10. Training Loss